\title{Evaluating and Crafting Datasets Effective for Deep Learning With Data Maps}

\author{
	\textbf{Bishnu, Jay}\\
	\and
	\textbf{Gondoputro, Andrew}}

\newcommand{\abstracttext}{\noindent
	Rapid development in deep learning model construction has prompted an increased need for appropriate training data. The popularity of large datasets - sometimes known as ``big data'' - has diverted attention from assessing their quality. Training on large datasets often requires excessive system resources and an infeasible amount of time. Furthermore, the supervised machine learning process has yet to be fully automated: for supervised learning, large datasets require more time for manually labeling samples. We propose a method of curating smaller datasets with comparable out-of-distribution model accuracy after an initial training session using an appropriate distribution of samples classified by how difficult it is for a model to learn from them. 
}

\documentclass[11pt, a4paper, twocolumn]{article}
\usepackage{xurl}
\usepackage{abstract}
\usepackage{graphicx}
\usepackage{lipsum}
\usepackage[backend=bibtex, style=authoryear, bibstyle=authoryear, citestyle=authoryear, natbib=true, sorting=none]{biblatex}
\usepackage{dblfnote}
\usepackage{hyperref}
\usepackage{array}
\usepackage{longtable}

\DeclareLanguageMapping{american}{american-apa}

\DeclareCiteCommand{\cite}
{\usebibmacro{prenote}}
{\usebibmacro{citeindex}%
	\printtext[bibhyperref]{\usebibmacro{cite}}}
{\multicitedelim}
{\usebibmacro{postnote}}

\DeclareCiteCommand*{\cite}
{\usebibmacro{prenote}}
{\usebibmacro{citeindex}%
	\printtext[bibhyperref]{\usebibmacro{citeyear}}}
{\multicitedelim}
{\usebibmacro{postnote}}

\DeclareCiteCommand{\parencite}[\mkbibparens]
{\usebibmacro{prenote}}
{\usebibmacro{citeindex}%
	\printtext[bibhyperref]{\usebibmacro{cite}}}
{\multicitedelim}
{\usebibmacro{postnote}}

\DeclareCiteCommand*{\parencite}[\mkbibparens]
{\usebibmacro{prenote}}
{\usebibmacro{citeindex}%
	\printtext[bibhyperref]{\usebibmacro{citeyear}}}
{\multicitedelim}
{\usebibmacro{postnote}}

\DeclareCiteCommand{\footcite}[\mkbibfootnote]
{\usebibmacro{prenote}}
{\usebibmacro{citeindex}%
	\printtext[bibhyperref]{ \usebibmacro{cite}}}
{\multicitedelim}
{\usebibmacro{postnote}}

\DeclareCiteCommand{\footcitetext}[\mkbibfootnotetext]
{\usebibmacro{prenote}}
{\usebibmacro{citeindex}%
	\printtext[bibhyperref]{\usebibmacro{cite}}}
{\multicitedelim}
{\usebibmacro{postnote}}

\DeclareCiteCommand{\textcite}
{\boolfalse{cbx:parens}}
{\usebibmacro{citeindex}%
	\printtext[bibhyperref]{\usebibmacro{textcite}}}
{\ifbool{cbx:parens}
	{\bibcloseparen\global\boolfalse{cbx:parens}}
	{}%
	\multicitedelim}
{\usebibmacro{textcite:postnote}}

\hypersetup{colorlinks=true, urlcolor=blue, linkcolor=blue, citecolor=blue}
\begin{document}

	\twocolumn[
	\begin{@twocolumnfalse}
		\maketitle
		\begin{abstract}
			\abstracttext
			\newline
			\newline
		\end{abstract}
	\end{@twocolumnfalse}
	]

	\section{Introduction}
	
	Though larger datasets are often preferred for deep learning to maximize out-of-distribution accuracy, their benefits begin to diminish at a particular size (\cite{banko-brill-2001-mitigating}). A dataset's efficiency can be defined qualitatively as the accuracy of a model trained on it relative to the amount of time and resources needed to produce the model. Our work utilizes the PyTorch Dataset Cartography project (\cite{swayamdipta2020dataset}) to evaluate metrics of individual samples (dubbed ``training dynamics'') in a dataset after training. By using training dynamics to represent the correctness, confidence, and variability in a model's prediction for a sample's label, we can categorize samples by their difficulty for the model to learn. With groups of these samples, we can craft a new data subset. Ideally, a data subset constructed from a certain proportion of hard-to-learn, easy-to-learn, and ambiguous samples could achieve comparable real-world performance while requiring a fraction of the memory and training time.
	
	\section{Background}
	
	\subsection{Datasets}
	
	The Dataset Cartography project is designed for use with several GLUE-style datasets. For this project, we used the Stanford Natural Language Inference (SNLI) Corpus (\cite{snli:emnlp2015}). All GLUE-style datasets are stored in the TSV (tab-separated value) format.
	
	The SNLI dataset contains two statements per sample. During model production, each pair is assigned a gold label\footnote{A label declared accurate, usually by a human.} - entailment, contradiction, or neutral - based on how the second statement relates to the first. The model is trained to predict the label of a new sentence pair using the train split\footnote{A portion of the entire dataset.} of the data. The model's out-of-distribution accuracy is measured by predicting labels for samples in the out-of-distribution dev split before returning to training. Once training has concluded, the model's performance is evaluated on the out-of-distribution test split.
	
	\begin{table*}[!ht]
		\centering
		\begin{tabular}{| m{5.75cm} | m{5.75cm}| m{2.5cm} |}
			\hline
			Sentence 1                                                                & Sentence 2                                       & Gold Label    \\ \hline
			Three black dogs on grass.                                                & Three dogs on grass.                             & Entailment    \\ \hline
			Three men are sitting on chairs.                                          & The men are dancing together on the dance floor. & Contradiction \\ \hline
			Short dark hair woman with a black phone is on the phone on the sidewalk. & A woman is on the phone with her husband.        & Neutral       \\
			\hline 
		\end{tabular}
		\caption{Samples from the train split of the SNLI dataset.}
	\end{table*}
	
	\subsection{Data Maps}
	
	Manually estimating the effectiveness of each sample in a dataset for training can be costly and time-consuming. The Dataset Cartography project was first proposed as a way to characterize samples in a dataset through a scatter plot. Following model evaluation, the samples in the model’s training sequence are plotted according to their training dynamics. The y-axis represents a model’s confidence in its prediction for the sample, and the x-axis represents the variability in its confidence. These metrics can allow humans and computers to identify from which samples the model learns easily and from those it struggles with.
	
	The Dataset Cartography project records new metrics after every epoch of training. These metrics are used to label every example in a training split with its correctness, confidence, and variability. Correctness represents the fraction of epochs in which the model correctly predicted the label for a given sample. Confidence is a numeric metric between 0 and 1 that measures the mean of the model’s probabilities of a sample being labeled correctly across epochs. Variability is the standard deviation of these probabilities. 
	
	Based on the confidence and variability of a data point, the project classifies it into one of three categories: easy-to-learn, hard-to-learn, or ambiguous. Easy-to-learn samples generally have low variability and high confidence, hard-to-learn samples have low variability and low confidence, and ambiguous samples have high variability. 
	
	Labeling data points in this way allows for data subsets to be filtered from the original dataset based on specific criteria; for example, the samples the model found easiest to learn from can be described as those with the highest confidence. The project allows for a selection of samples - i.e., the 33\% of samples with the highest confidence - to be filtered into a new data subset. Our method uses this feature to create new data subsets.
	
	The original paper introducing the Dataset Cartography project describes several out-of-dataset accuracy tests. The authors evaluated the accuracy of models trained on the top 33\% easiest-to-learn, hardest-to-learn, and most ambiguous samples. The model trained on the ambiguous samples predicted the correct label for both in-distribution and out-of-distribution data most consistently.
	
	However, further evaluations revealed that small ambiguous datasets performed worse. This phenomenon could indicate that ambiguous samples alone were insufficient for learning and that there may be a combination of sample types with superior performance. 
	
	\begin{figure}[!ht]
		\includegraphics[scale=0.75]{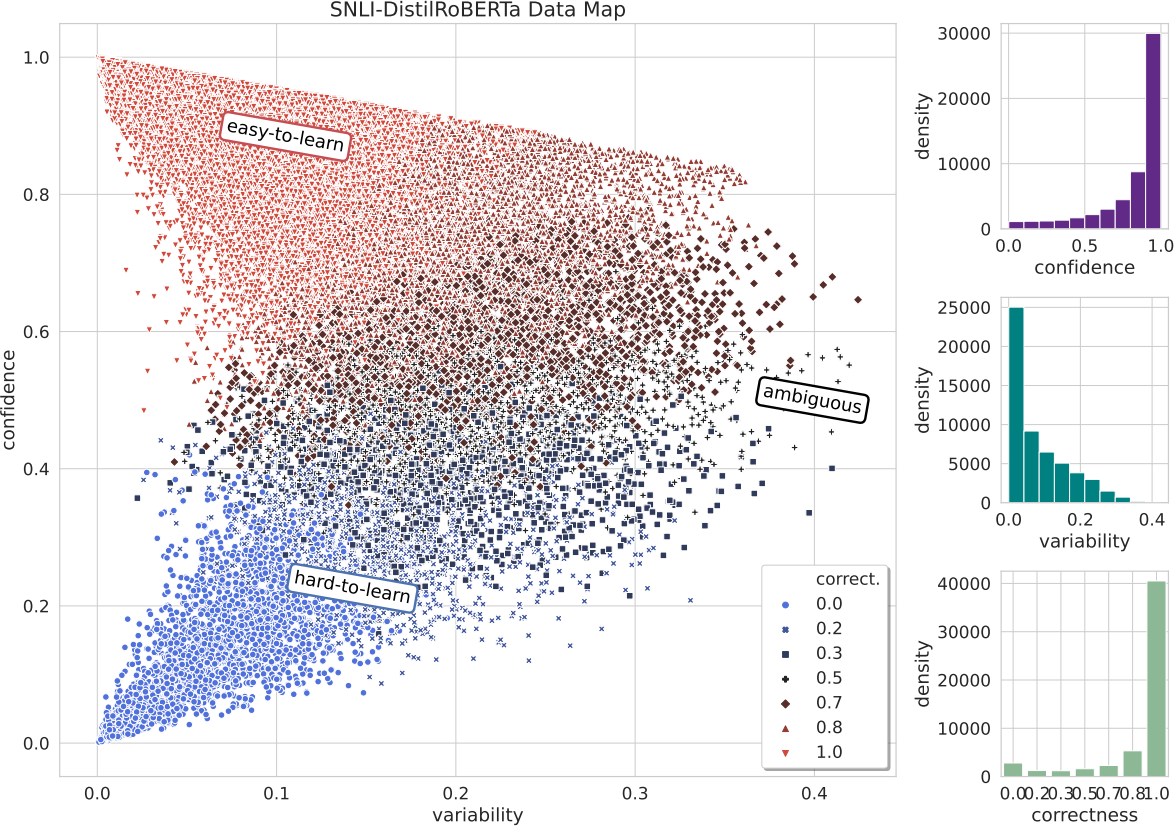}\caption{A data map for a DistilRoBERTa-base model trained on the SNLI dataset without shuffling.}
	\end{figure}
	
	\section{Methods}
	
	Our goal was to expand upon the training analysis from the original Dataset Cartography project paper. We used the DistilRoBERTa-base transformers model and the SNLI corpus for training and evaluation.
	
	We trained a preliminary model on the entire dataset for six epochs to classify samples as easy-to-learn, hard-to-learn, or ambiguous. Using the filtering function, we sorted by the confidence metric to filter by the hardest-to-learn and easiest-to-learn samples, and by the variability metric to filter by the most ambiguous samples. We then created nine new subsets after filtering the labeled data: a copy of the original dataset, but with shuffled samples; a randomly selected 33\% sample of the original dataset; three 33\% datasets, or one for the top 33\% of each of the categories; three 16.67\% + 16.67\% datasets, with the three possible combinations of the categories (i.e., 16.67\% easy-to-learn and 16.67\% hard-to-learn); and a dataset containing the top 11\% of samples from all three categories.
	
	\section{Findings}
	
	\subsection{Results}
	
	After training all nine models on the new datasets, we recorded their accuracy in labeling the in-distribution train set and the out-of-distribution dev and test sets.

	The model trained on the entire SNLI dataset had the highest out-of-distribution accuracies, while the model trained on easiest-to-learn samples had the highest in-distribution accuracy. The model trained on only the hardest-to-learn and most ambiguous samples had the lowest out-of-distribution accuracy, and the model trained on the hardest-to-learn samples had the lowest in-distribution accuracy. 

	We found that the model trained on only the most ambiguous samples had comparable out-of-distribution accuracies to the entire dataset's, but poor in-distribution accuracy. Excluding the model trained on the entire dataset, the model with the highest overall accuracies was the one trained on the easiest-to-learn and most ambiguous samples. 
	
	\begin{table*}[!ht]
		\centering
		\begin{tabular}{| m{4.4cm} | m{3.2cm}| m{3.2cm} | m{3.2cm} |}
			\hline
			~                                                                           & Final Training Accuracy (ID) & Dev \newline Accuracy (OOD) & Test \newline Accuracy (OOD) \\ \hline
			100.00\%                                                                    & 0.8936                       & 0.8997                      & 0.8976                       \\ \hline
			33.33\% random                                                              & 0.8782                       & 0.8776                      & 0.8785                       \\ \hline
			33.33\% easy-to-learn                                                       & 0.9996                       & 0.8293                      & 0.8286                       \\ \hline
			33.33\% hard-to-learn                                                       & 0.5680                       & 0.5966                      & 0.5856                       \\ \hline
			33.33\% ambiguous                                                           & 0.7684                       & 0.8878                      & 0.8894                       \\ \hline
			16.67\% easy-to-learn\newline16.67\% hard-to-learn                          & 0.7581                       & 0.5938                      & 0.5999                       \\ \hline
			16.67\% easy-to-learn\newline16.67\% ambiguous                              & 0.8835                       & 0.8802                      & 0.8807                       \\ \hline
			16.67\% hard-to-learn\newline16.67\% ambiguous                              & 0.5900                       & 0.4076                      & 0.4023                       \\ \hline
			11.11\% easy-to-learn\newline11.11\% hard-to-learn\newline11.11\% ambiguous & 0.7401                       & 0.5249                      & 0.5145                       \\ \hline
		\end{tabular}
		\caption{Results from training and testing models based on the data subsets.}
	\end{table*}
	
	\subsection{Mislabeled Data}
	
	The original Dataset Cartography paper discussed the idea of noisy data (mislabeled samples), which could be more easily identified by filtering hard-to-learn samples. The SNLI Dataset classifies pairs of sentences as either entailment, neutral, or contradiction, but some samples with inaccurate gold labels could interfere with training and evaluation. While exploring some of the samples categorized as hard-to-learn, we found several mislabeled data points. We strongly suspect these were partially responsible for the poor performance of the models trained on subsets with the hardest-to-learn samples.
	
	\begin{table*}[!ht]
		\centering
		\begin{tabular}{| m{4.5cm} | m{4.5cm}| m{2.5cm} | m{2.5cm} |}
			\hline
			Sentence 1                                                             & Sentence 2                                        & Gold Label    & Our Label     \\ \hline
			A hiker travels along a rocky path bordered by greenery.               & The hiker is carrying a big backpack on his back. & Entailment    & Neutral       \\ \hline 
			Two beige dogs are playing in the snow.                                & Two dogs are outside.                             & Contradiction & Entailment    \\ \hline
			A man with a gray beard is standing next to a woman in a brown jacket. & Nobody is standing.                               & Neutral       & Contradiction \\
			\hline
		\end{tabular}
		\caption{Mislabeled samples in the SNLI train split.}
	\end{table*}
	
	\section{Conclusion}
	
	Our research demonstrated that a model trained on only specific samples in a dataset could achieve accuracy comparable to one trained on the entire dataset. In particular, models trained on subsets with a strong focus on ambiguous samples and little-to-no focus on hard-to-learn samples generally had excellent out-of-distribution performance, while those trained on easy-to-learn samples had near-perfect in-distribution performance. We believe that evaluating and refining a dataset based on the training dynamics of its samples can allow for smaller, more efficient models with equivalent - or potentially even superior - performance. 
	
	\section{Further Research}
	
	The Dataset Cartography project has other applications in dataset and model construction. Not only can it be used to identify mislabeled or low-quality samples, but the filtering process could also be automated further to find an ideal ratio of easy-to-learn, hard-to-learn, and ambiguous samples for any particular dataset. Though this proof-of-concept project is currently constrained to a select few datasets, the method can be generalized to fields beyond natural language processing and categorization projects. A similar process could allow model engineers to evaluate the quality of data used to train generative adversarial networks (GANs) by analyzing the quality of generated samples used by the GAN's corresponding discriminator (\cite{1406.2661}).
	
	\section{Acknowledgments}
	
	We are grateful to have performed our research under the guidance of Mark Galassi, Rhonda Crespo, Maria de Hoyos, and the rest of the Institute for Computing in Research team. Furthermore, the Software Freedom Conservancy generously funded all of our research. We would also like to thank Dr. Ameeta Agrawal and her Ph.D. student Yufei Tao at Portland State University for spearheading our project. Finally, we thank the Allen Institute for Artificial Intelligence for the original Dataset Cartography project and accompanying research.

	\nocite{*}
	
	\printbibliography[title={Bibliography}]
	
\end{document}